\documentclass{article}
\usepackage{spconf,amsmath,graphicx}
\usepackage{multirow}
\usepackage{comment}
\usepackage{subcaption} 
\usepackage{threeparttable}
\usepackage{booktabs}
\usepackage{pifont}
\usepackage{amssymb}
\usepackage{color,soul}
\usepackage{hyperref}

\newcommand{\xmark}{\ding{55}}

\usepackage{xspace}
\makeatletter
\DeclareRobustCommand\onedot{\futurelet\@let@token\@onedot}
\def\@onedot{\ifx\@let@token.\else.\null\fi\xspace}

\def\ie{\emph{i.e}\onedot}

\makeatother

\title{Multiscale Crowd Counting and Localization by Multitask\\Point Supervision}

\name{Mohsen Zand, Haleh Damirchi, Andrew Farley, Mahdiyar Molahasani, Michael Greenspan, Ali Etemad
}
\address{Dept. ECE \&
Ingenuity Labs Research Institute, Queen's University,
Kingston, Canada}

\begin{document}

\maketitle
\begin{abstract}
We propose a multitask approach for crowd counting and person localization in a unified framework. As the detection and localization tasks are well-correlated and can be jointly tackled, our model benefits from a multitask solution by learning multiscale representations of encoded crowd images, and subsequently fusing them. In contrast to the relatively more popular density-based methods, our model uses point supervision to allow for crowd locations to be accurately identified. We test our model on two popular crowd counting datasets, ShanghaiTech A and B, and demonstrate that our method achieves strong results on both counting and localization tasks, with MSE measures of 110.7 and 15.0 for crowd counting and AP measures of 0.71 and 0.75 for localization, on ShanghaiTech A and B respectively. Our detailed ablation experiments show the impact of our multiscale approach as well as the effectiveness of the fusion module embedded in our network. Our code is available at:  \href{https://github.com/RCVLab-AiimLab/crowd_counting}{https://github.com/RCVLab-AiimLab/crowd\_counting}
\end{abstract}

\section{Introduction}
\label{sec:intro}
Crowd counting is central to many real-world computer vision applications, including crowd management~\cite{liu2019point, laradji2018blobs}, surveillance systems~\cite{wang2019object}, security and planning~\cite{chan2008privacy}, traffic monitoring~\cite{chen2010real}, animal crowd estimation~\cite{marsden2018people}, and cell counting~\cite{wang2016fast}. In general, crowd counting is a challenging problem due to scale variations, occlusions, complex and noisy backgrounds, and variations in perspective and illumination. This problem has been addressed most prominently by estimated density maps~\cite{zhang2015cross}, wherein annotated head locations are converted to a density map through convolution with a Gaussian kernel, following which integration over the map generates the people count~\cite{zhang2016single}. Utilizing density maps is often considered to be the standard approach towards crowd counting ~\cite{li2018csrnet, liu2019recurrent}.

Recently, promising performances have been achieved by employing convolutional neural networks (CNNs) for crowd counting via density map estimation~\cite{li2018csrnet, gao2020cnn}. Density estimators are however highly sensitive to the choice of the kernel and the kernel size used to generate them. More importantly, the use of density maps results in inconsistent performance with varying crowd sparsities, which in the past has been addressed by CSRNet~\cite{li2018csrnet} through expanding the receptive fields of the network. In addition, density-based methods only provide an estimate of people count, and thus fail to capture individual information such as person location and size~\cite{liu2019point, wang2021self}. Such attributes may be important in many other downstream applications, for instance multi-object tracking, person re-identification, and face recognition.

In this paper, we propose the use of multiscale point supervision~\cite{laradji2018blobs} to improve the performance of crowd counting in both densely and sparsely populated crowd scenes. Moreover, we employ a multitask approach, which can simultaneously perform localization through exploiting the scene representations learned in the intermediate layers. Our method does not require density maps to be generated and is thus better equipped to deal with varying sparsities in crowd scenes. Our experiments on two large public datasets, ShanghaiTech A and ShanghaiTech B, demonstrate the effectiveness of our proposed method and robustness against varying sparsities. 

Our contributions in this paper include the following two aspects:
(\textbf{1}) We propose a novel multiscale and multitask architecture based on point supervision to effectively estimates both count and location under large variations in the number of people per image. 
(\textbf{2}) Our method achieves strong results and approaches the state-of-the-art on two datasets for both counting and localization. Detailed ablation experiments demonstrate the impact of each component in our network. 



\section{Related work}
\label{sec:related_work}

Crowd counting methods can be divided into the two major categories of density-based and point-based~\cite{loy2013crowd}. In the following we review the related works in these two categories.

\begin{figure*}[t]
\begin{center}
\includegraphics[width=0.8\linewidth ]{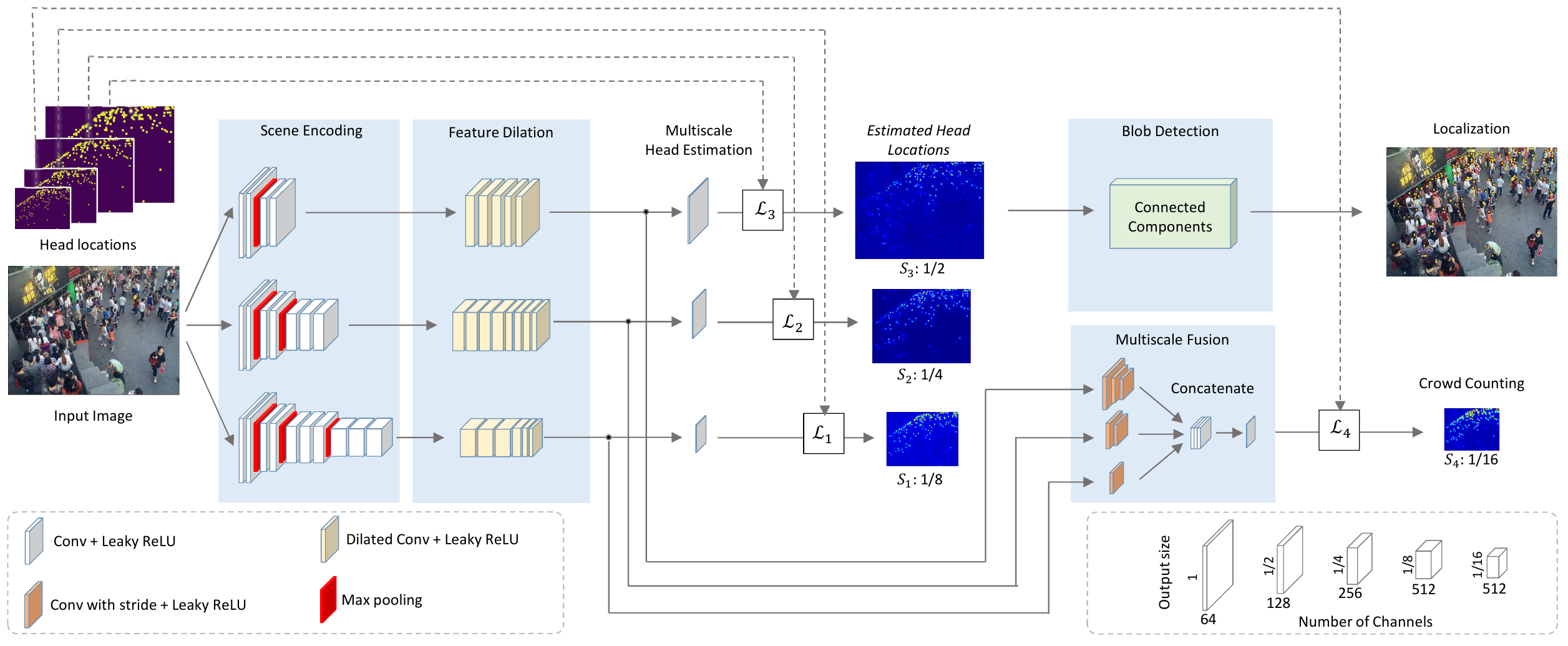}
\end{center}
\vspace{-0.5 cm}
   \caption{The architecture of the proposed model. Dashed lines are used solely in the training stage.}
\label{fig:model}
\end{figure*}

\noindent \textbf{Density-based.} Most solutions rely on probability maps to regress the density of the crowd~\cite{li2018csrnet, sam2017switiching}. These approaches utilize binary head locations that have been blurred by a Gaussian kernel to form the ground truth density maps as the estimate of the number of people in the crowd. As an example of such approaches, in~\cite{li2018csrnet}, dilated convolutions were used to take advantage of a large receptive field to obtain strong crowd image representations. 

To solve the problem of head size variations in crowd images, the use of several branches of deep network has been proposed. For instance, in~\cite{sam2017switiching}, a switching network was used alongside three regressor networks. Each regressor had a different architecture for different rates of scale and density. The switching network learned to direct the images with high density towards the regressor with a smaller receptive field. In~\cite{boominathan2016crowdnet}, an architecture with two parallel branches of CNNs, one shallow and one deep, was proposed to tackle the problem of scale and perspective variance in crowd images. In~\cite{zeng2017multi}, an inception-like network was used to generate feature maps of different sizes, while in~\cite{zhang2016single}, a CNN with three branches was proposed, each with a different receptive field. The problem of scale-variant head sizes was also addressed and tackled in~\cite{zhang2015cross} in which the training images were divided into patches, and categorized by density. After training the CNN, fine-tuning for each test image was carried out using the patches of training images with the same density rate. A stacked pooling approach was later proposed in~\cite{huang2020stacked} to solve the problem of counting variable head sizes in crowd images.

\noindent \textbf{Point-based.} These methods use crowd head annotations either directly as ground truth or as a means to generate bounding boxes around heads. This approach has the advantage of being able to predict the location of heads in the crowd in addition to counting the number of people present in the image. The method presented in~\cite{laradji2018blobs} proposed an extension of the semantic segmentation loss in~\cite{long2015fully} for point supervision based localization. In~\cite{liu2019point}, bounding boxes around point annotations were initialized using their distance to the nearest neighbour head and while training a CNN to output box annotations, updated bounding boxes around the heads to obtain the most suitable size in the anchor box set for the corresponding head. In~\cite{sam2020going}, two scales of the input image feature maps (one-fourth and one-eighth) were used depending on the sparsity of the crowd to output a point map.
Crowd counting was also cast as a classification task of dot prediction with point supervision, dropping the prevalent density regression. 
An adaptive scale fusion module combined the multiscale confidence maps into a single map, where each value indicated the confidence of person detection. A threshold was then applied on this map to generate the final accurate dot predictions. Therefore in~\cite{sam2020going}, the fusion module was not learnable and the count highly depended on the accuracy of the threshold value which might be different across different images. 




\section{Proposed method}
\label{sec:method}

Let $P=\{p_i\}_{i=1}^M$ be the head coordinates of each person $i$, where $p_i=(x_i,y_i)$, and $M$ is the total number of people in the image.
Our goal is to estimate $M$ and $P$ through a point-supervised multiscale unified neural framework. Our proposed model is described as follows (see Fig.~\ref{fig:model}).

\noindent\textbf{Multiscale Scene Encoder.}
An arbitrary-sized image $I$ is fed to three VGG16-based scene encoders, pre-trained on ImageNet \cite{deng2009imagenet}, to obtain representations in three scales, \ie $S_1=1/8$, $S_2=1/4$, and $S_3=1/2$ of the original image size. We thus obtain three separate embeddings corresponding to the three respective scales, which ultimately improves our point-supervised network's ability to estimate $M$ and $P$ accurately for a large variation of the number of people in a scene (i.e. crowd spatial distribution).

\noindent \textbf{Feature Dilation.} 
The multiscale scene representations are then fed to dilated convolutional layers which are proven useful in previous works~\cite{li2018csrnet, wang2021self}. They serve to extend the layers' receptive fields and capture higher level features. Specifically, the dilated CNNs have the benefit of exploiting various receptive field sizes from the original image, without degrading the resolution as may occur when downsampling by increasing convolutional kernel size. The number of layers however are different in these networks. To achieve consistent convergence between the different scales of our pipeline, we adjust the number of layers in different branches.

\noindent \textbf{Multiscale Head Estimation.} 
Each embedding extracted by the feature dilation is fed to a single layer to generate an estimate for the head locations in the scene. This is achieved using a one-channel convolution network composed of a convolution layer with a kernel size of $1\times 1$. This structure serves to improve the accuracy of the overall localization results, as well as generating inputs to the multitasking network to support accurate crowd counting. The heatmaps of these embeddings are visualized in Fig.~\ref{fig:model} for each scale.

To obtain the estimated head locations, we need the ground-truth head points, which are given in the dataset. Nevertheless, these locations must be normalized to be used at different scales. For each point, we extract the head coordinates from the head location and normalize the $x$- and $y$-values in the range of $[0,1]$ by dividing them by the image width and height, respectively. We then multiply the normalized head locations by the three scale factors to obtain the resized ground-truth head locations.

\noindent \textbf{Multiscale Fusion.} 
This module generates the final density map for crowd counting. It consists of three networks with different number of convolutional layers with stride of 2 and kernel size of 2, each of which works on one of the dilated embeddings. The first network comprises one layer and works on the $S_1$ scale. Similarly, the second and third networks respectively work on the $S_2$ and $S_3$ scales, and include two and three layers. These networks specifically downsample the dilated embeddings at scales $S_1$, $S_2$, and $S_3$, respectively, and generate the same-sized outputs as they use different number of convolutional layers. These outputs are then concatenated channel-wise and fed to a convolutional layer to generate the final density map of $S_4=1/16$.

\noindent\textbf{Localization}. 
Given that the resolution of $S_3$ is higher than the other two branches, the head \textit{locations} can be extracted from this branch more precisely. Accordingly, we use this embedding to perform localization by utilizing a connected components algorithm
~\cite{wu2005optimizing} to obtain the blobs in the scenes. The center of the blobs represent the head locations in the crowd image. 
Fig.~\ref{fig:model} illustrates the detected blobs.

\noindent \textbf{Total Loss.}
We use four MSE loss terms $\mathcal{L}_j=\| \widehat{D}_j - D_j\|_2^2$ for the three multiscale branches and the multiscale fusion network, where $\widehat{D}_j$ and $D_j$ denote
the estimated map at scale $S_j$ (consisting of head locations at this scale) and its ground-truths,
respectively. 
Specifically, $D_j$ represents the embedding at scale $S_j$. Each location $(x^{S_j}_k$, $y^{S_j}_k)$ in $D_j$ shows the integrated number of people which their original coordinates $(x_i,y_i)$ map to. For example, each $16\times16$ pixel block in the original image corresponds to one location on the embedding of scale $S_4$, and the total number of people $p_i$ with $0< x_i \le 15$ and $0< y_i \le 15$ assign to $(0,0)$ on $D_j$. 
Ideally, all outputs generated from different branches would correspond to the same number of people. This however does not hold in practice, as the different branches of the network will have varying accuracies at detecting heads of different sizes (likely due to varying distances from the camera)~\cite{sam2020locate,chan2008privacy,yao2020real}. To address this problem, we use task-specific weights $w_j$ in the final loss function, and obtain $\mathcal{L}_{total}=\sum_{j=1}^4 w_j \mathcal{L}_j$. This approach also helps obtain a more consistent convergence when training our end-to-end model as different branches could have the tendency to learn at different rates.


\section{Experiments and Results}
\label{sec:experiments}
\subsection{Experimental Setup}
\textbf{Datasets.}
We use two popular datasets, ShanghaiTech A and ShanghaiTech B ~\cite{zhang2016single}, which are the most commonly used datasets in the area of crowd counting \cite{zhang2015cross,li2018csrnet,laradji2018blobs,liu2019recurrent,sam2020going,huang2020stacked}.
ShanghaiTech A contains 482 images with 241,677 total annotated heads. ShanghaiTech B has 716 images with 88,488 total annotated heads ~\cite{zhang2016single}. Part A is representing more crowded scenes while part B includes more sparse images. 
We compensate for the relatively small number of images in the standard ShanghaiTech datasets by augmenting the images similar to \cite{li2018csrnet}. We also first pretrain the scene encoders in our network with the much larger ImageNet dataset~\cite{deng2009imagenet}.

\noindent
\textbf{Implementation Details.}
Adam optimizer was used with a momentum of 0.934 and an initial learning rate of 1E-6. The model was trained for 200 epochs with early stopping, and the task-specific weights were $w_1 = 0.1$, $w_2 = 0.2$, $w_3 = 0.3$, $w_4 = 0.1$ for both datasets. Training was performed with an Nvidia Titan RTX GPU. The architectural details are provided in Fig.~\ref{fig:model}, where kernel sizes of 3, 1, and 2 have been used for the scene encoding, head estimation, and fusion, respectively. A stride of 1 is used throughout all the convolutional layers, except for the fusion network which uses a stride of 2. 

\noindent\textbf{Evaluation.}
The summation of the output of the multiscale fusion corresponds to the total number of people in the scene. 
The mean absolute error (MAE) and mean square error (MSE) of the predicted count with respect to the ground truths are calculated. To evaluate the localization results we use average precision (AP) which is the area under the precision-recall curve. 
If a detected head is within 5 pixels of the ground truth head location, that detection is denoted as a true positive, and is deleted from the ground truth points so that it would not be matched with any other prediction in the future assessments. If a detection it is not within that distance of a head detection, it is counted as a false positive, and finally, if a head detection in ground truth is not matched with any detection, then it is counted as a false negative.

\begin{table}[t!]
\centering
\caption{MAE, MSE, and AP scores on the ShanghaiTech datasets.} 
\small
\begin{tabular}{l|cc|c|cc|c}
\toprule   
  \textbf{Dataset} & \multicolumn{3}{c|}{\textbf{ShanghaiTech A}} & \multicolumn{3}{c}{\textbf{ShanghaiTech B}}\\ 
  & \multicolumn{2}{c|}{\textbf{\textit{Counting}}} & \textbf{\textit{Loc.}} & \multicolumn{2}{c|}{\textbf{\textit{Counting}}} & \textbf{\textit{Loc.}} \\ 
   & \textit{MAE}$\downarrow$ & \textit{MSE}$\downarrow$ & \textit{AP}$\uparrow$ & \textit{MAE}$\downarrow$ & \textit{MSE}$\downarrow$ & \textit{AP}$\uparrow$\\ 
\midrule
   Cross-scene~\cite{zhang2015cross} & 181.1 & 277.7 & - & 32.0 & 49.8 & - \\ 
   MCNN~\cite{zhang2016single} & 110.2 & 173.2 & - & 26.4 & 41.3 & - \\
  LC-ResFCN~\cite{laradji2018blobs} & - & - & - & 25.89 & - & - \\
  LC-PSPNet~\cite{laradji2018blobs} & - & - & - & 21.61 & - & - \\
  RAZNet~\cite{liu2019recurrent}  & 75.2 & 133.0 & 0.69 & 13.5 & 25.4 & 0.69 \\
  RAZNet+~\cite{liu2019recurrent} & 71.6 & 120.1 & 0.69 & 9.9 & 15.6 & 0.71 \\
  DD-CNN~\cite{sam2020going} & 71.9 & 111.2 & 0.65 & - & - & - \\
  Deep-Stacked~\cite{huang2020stacked} & 94.0 & 150.6 & - & 18.7 & 31.9 & - \\
  CSRNet~\cite{li2018csrnet} & \textbf{68.2} & 115.0 & - & 10.6 & 16 & - \\
  \textbf{Ours} & 71.4 & \textbf{110.7} & \textbf{0.71} & \textbf{9.6} & \textbf{15.0} & \textbf{0.75} \\

\bottomrule
\end{tabular}
\label{table:All_results}
\end{table}

\subsection{Results}

\textbf{Crowd Counting.}
The performance of the proposed model for crowd counting is compared with similar works in Table \ref{table:All_results}. We observe that for both datasets, training on auxiliary localization information in the multitask approach helps with learning more effective and precise scene representations, and thus our method achieves better or competitive results to the related works. Notably, in ShanghaiTech B which contains more sparse scenes, our point-supervised model outperforms recent density-based approaches such as \cite{zhang2016single} and \cite{li2018csrnet}. 
This can be attributed to the fact that density-based method in which the individual heads are not distinctly identified are generally more robust when dealing with denser scenes.

\begin{figure}[t]
    \centering
    \begin{subfigure}{0.31\linewidth}
        \centering
        \includegraphics[width=1.1\linewidth]{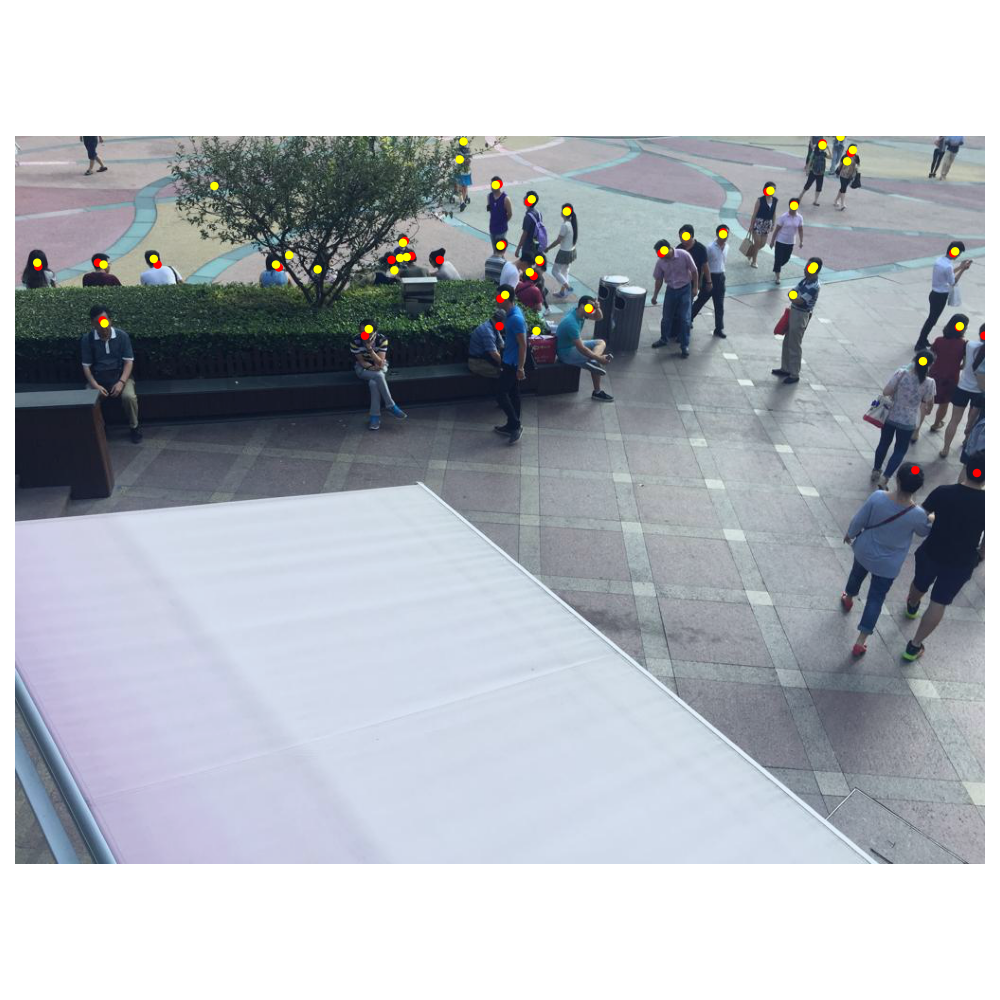}
        \label{fig:img_original}
    \end{subfigure}
    \hfill
    \begin{subfigure}{0.31\linewidth}
        \centering
        \includegraphics[width=1.1\linewidth]{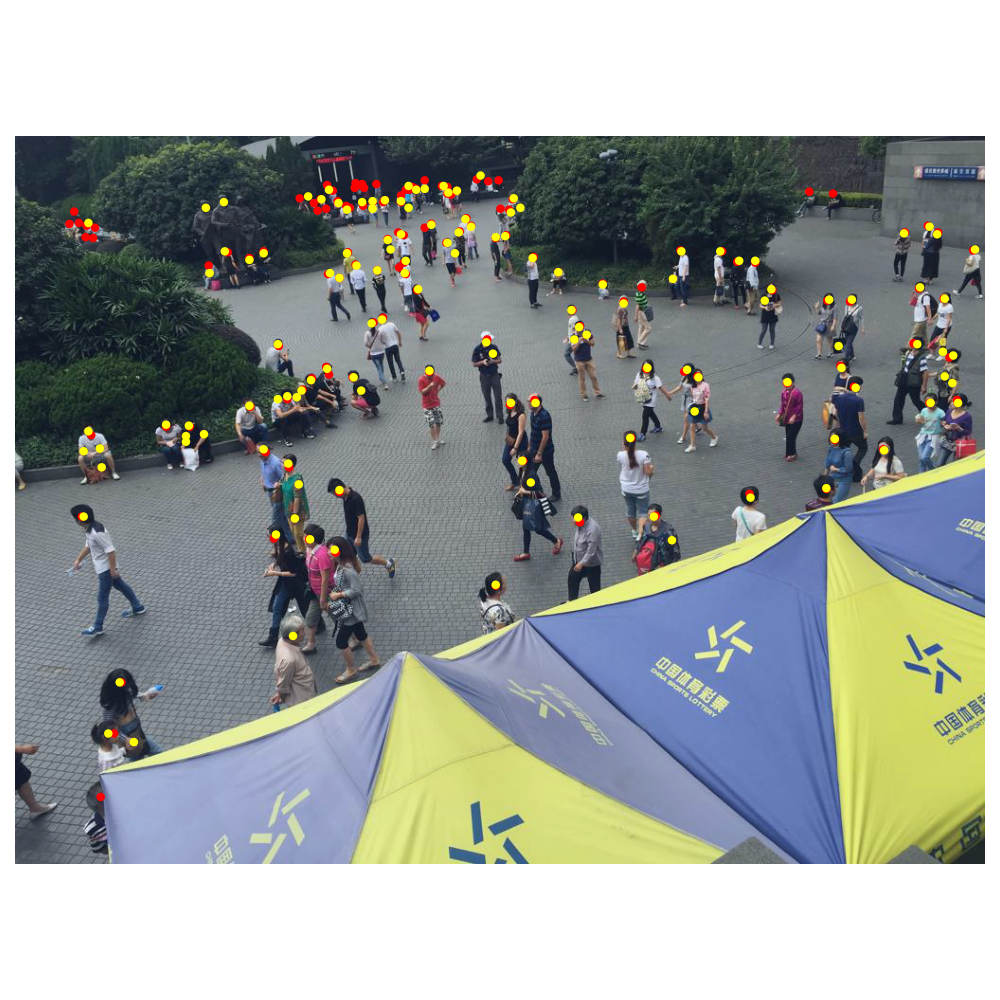}
        \label{fig:img_he}
    \end{subfigure}
    \hfill
    \begin{subfigure}{0.31\linewidth}
        \centering
        \includegraphics[width=1.1\linewidth]{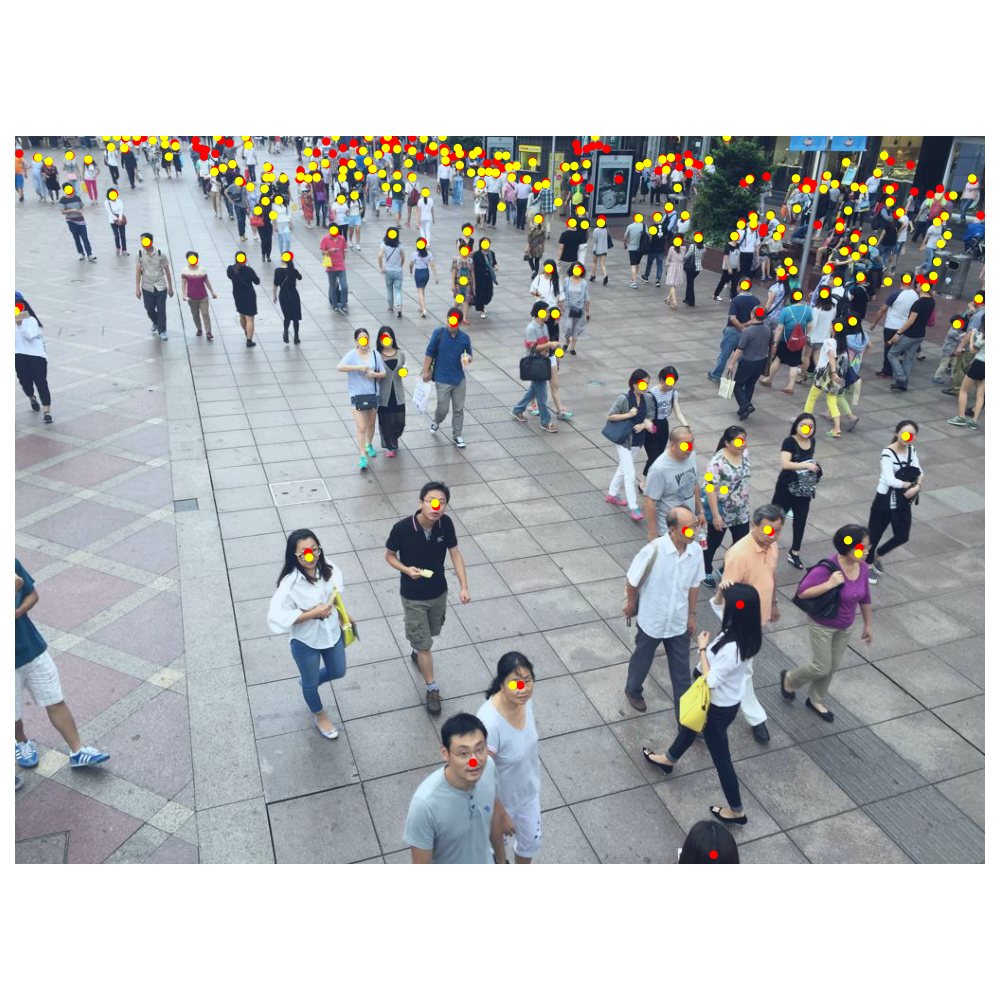}
        \label{fig:img_he}
    \end{subfigure}
\vspace{-1 cm}
\caption{Visualized samples of detection and localization. Yellow and red points denote detected and ground truth head locations, respectively.}
\label{fig:localization}
\end{figure}

\noindent \textbf{Localization Results.}
The evaluation results for localization of the detected heads are depicted in Table~\ref{table:All_results}. 
We observe that our point-supervised model delivers the best results in comparison to prior work on both datasets. It should be noted that the values for~\cite{liu2019recurrent} and~\cite{sam2020going} reported in our table are taken directly from the respective papers, which may have used slightly different definitions for true-positive detections given a lack of standard definition in the field.
Fig.~\ref{fig:localization} shows three sample images where the crowd and their locations have been identified, along with the corresponding ground truths. 
%

\noindent \textbf{Ablation Study.}
We aim to validate the multiscale aspect of our approach by systematically removing each scale branch through ablation experiments. We specifically remove $S_i$ by excluding their corresponding loss terms from our total loss. Each row in Table~\ref{tab:ablation_si} shows an experiment where crosses indicate the exclusion of the loss term for a particular scale. We observe that for both datasets, the performance degrades when any of the scales are removed.

\begin{figure}[t]
    \begin{subfigure}{0.49\linewidth}
        \centering
        \includegraphics[width=\linewidth]{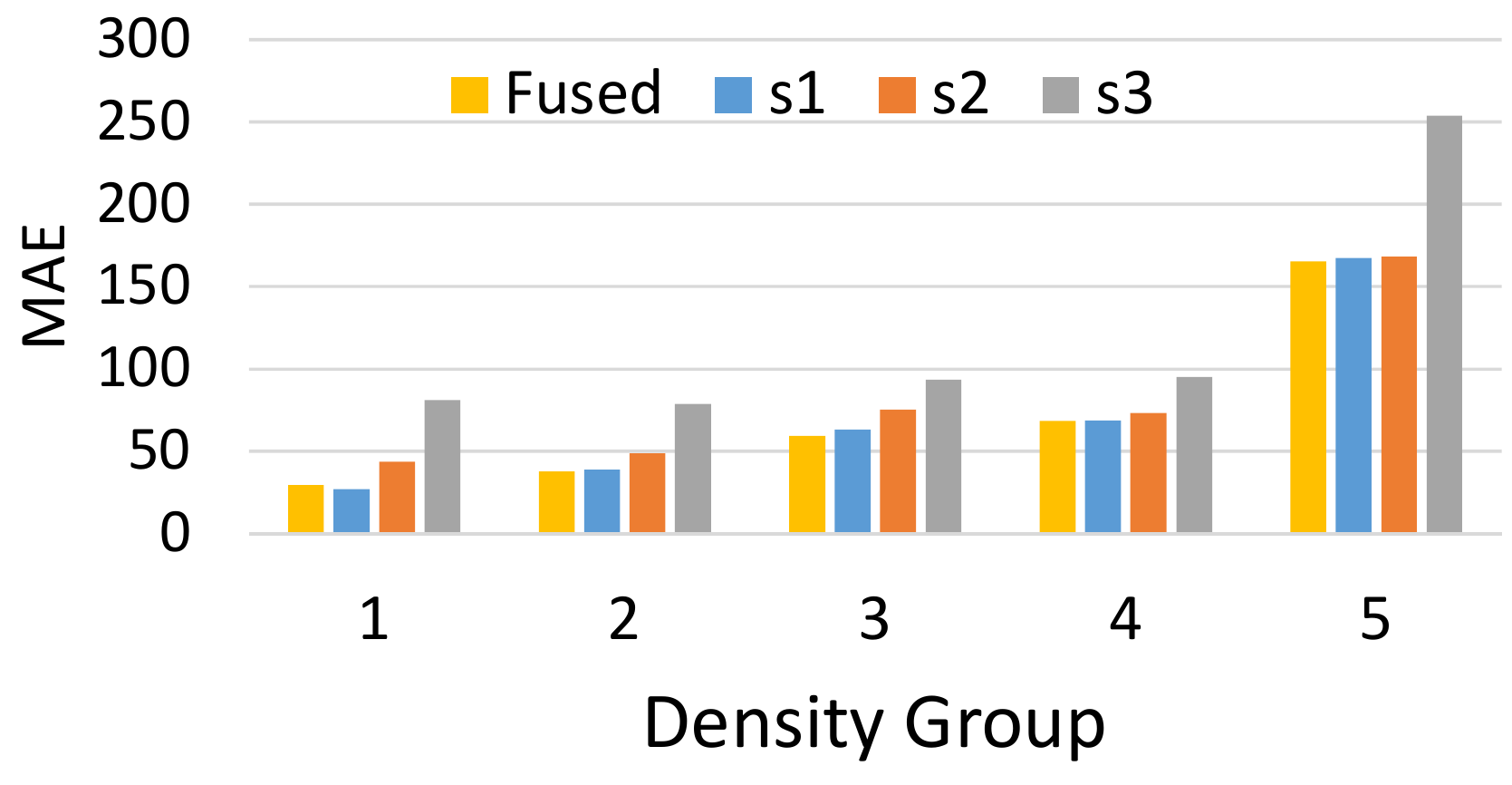}
        \label{fig:img_original}
    \end{subfigure}
    \hfill
    \begin{subfigure}{0.49\linewidth}
        \centering
        \includegraphics[width=\linewidth]{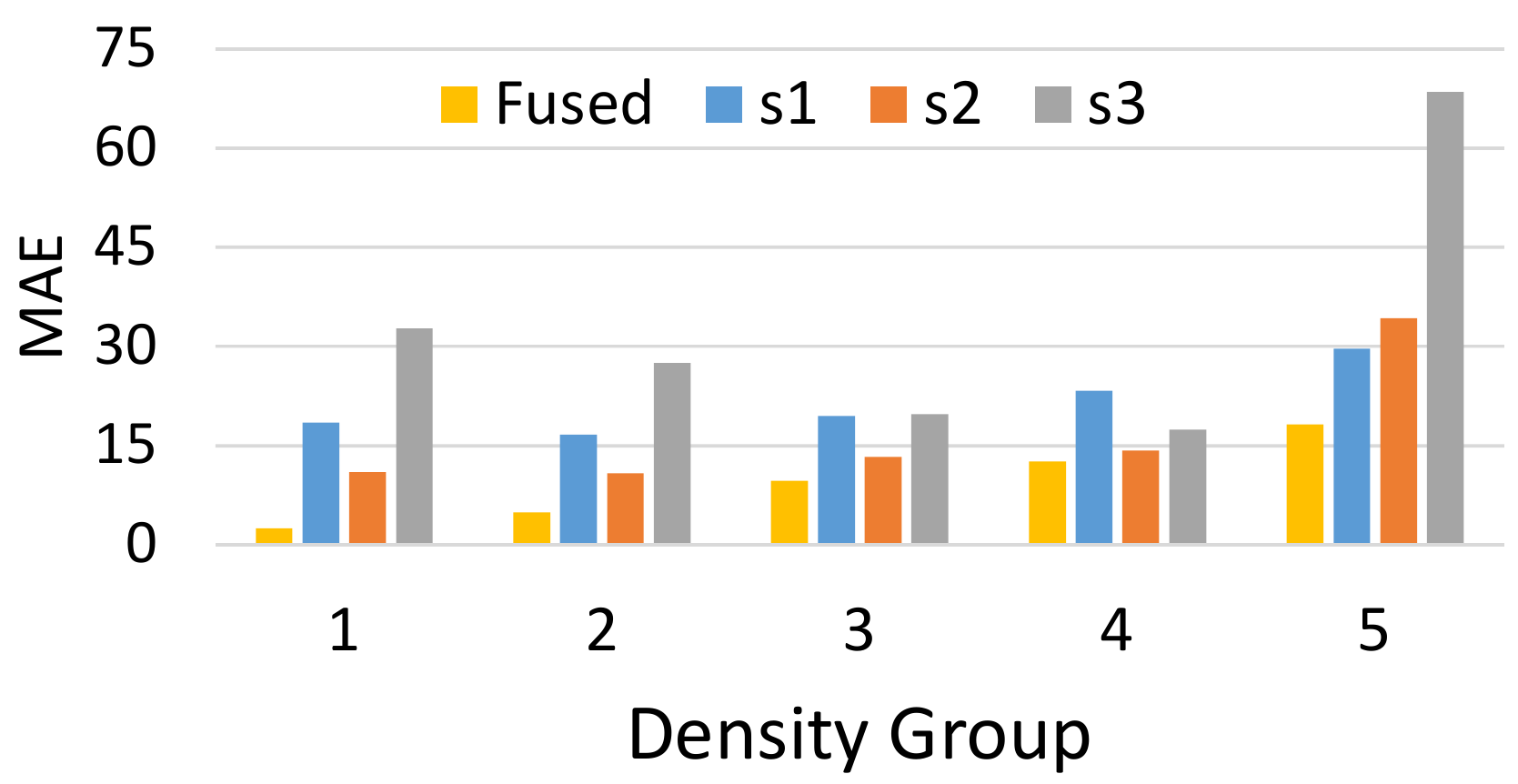}
        \label{fig:img_he}
    \end{subfigure}
\vspace{-0.7 cm}
\caption{The MAE for different crowd density groups in ShanghaiTech A (left) and ShanghaiTech B (right).}
\label{fig:density_Group}
\end{figure}

In order to show that the multiscale fusion module can effectively combine the information extracted from each branch in different densities we partition the images in the dataset into 5 different crowd density groups based on the number of people in each image, as in~\cite{huang2020stacked}. Consequently, Group 1 comprises the first $20\textsuperscript{th}$ percentile of density, Group 2 the second $20\textsuperscript{th}$ percentile, etc. Next, we calculate the MAE for each branch in order to investigate the impact of our multiscale fusion module in different density groups. The results are illustrated in Fig.~\ref{fig:density_Group}, where we observe that the fusion network successfully combines outputs from different scales in different densities. These results shows that the fusion network outperforms any single scale for almost every density group, which indicates that the fusion approach performs better than simply averaging the results of each individual scale. In particular, our multiscale fusion approach would yield equal or better performance than switching scales based on prior knowledge of these density groups.

\begin{table}[t!]
\centering
\caption{Ablation studies on different configurations for crowd counting on the ShanghaiTech datasets.} 
\small
\begin{tabular}{ccc|cc|cc}
\toprule   
   & & & \multicolumn{2}{c|}{ShanghaiTech A} & \multicolumn{2}{c}{ShanghaiTech B}  \\
  $\mathcal{L}_1$ & $\mathcal{L}_2$ & $\mathcal{L}_3$ & \textit{MAE} $\downarrow$ & \textit{MSE} $\downarrow$ & \textit{MAE} $\downarrow$ & \textit{MSE} $\downarrow$ \\ 

\midrule
  \xmark & \checkmark & \checkmark & 84.7 & 125.6 & 10.5 & 17.3 \\ 
  \checkmark & \xmark & \checkmark & 85.1 & 129.1 & 10.4 & 17.5 \\
  \checkmark & \checkmark & \xmark & 85.2 & 128.6 & 9.7 & 15.4 \\
  \checkmark & \checkmark & \checkmark & \textbf{71.4} & \textbf{110.7} & \textbf{9.6} & \textbf{15.0} \\


\bottomrule
\end{tabular}
\label{tab:ablation_si}
\end{table}


\section{Conclusion}
\label{sec:conclusion}
In this paper, we propose a novel multitasking deep neural network capable of performing both crowd counting and localization simultaneously using point supervision. Our model uses a multiscale architecture and an effective fusion module to deal with the different crowd densities that can occur in images. 
Our rigorous experiments demonstrate that our proposed model can achieve strong results in comparison to other works in the area on two popular datasets, ShanghaiTech A and B, for both crowd counting and localization tasks. Moreover, our ablation experiments showed the positive impact of the multiscale and fusion elements of our model.

For future work, we may explore avenues for the different terms in our loss function to be automatically weighted through learning. To this end, different attention mechanisms may be explored and integrated in our model. Additionally, using density maps alongside our point-based approach may be explored through ensemble or fusion approaches.

\noindent \textbf{Acknowledgements.} Thanks to Geotab Inc., the City of Kingston, and NSERC for their support of this work.


\end{document}